\documentclass[11pt,a4paper]{article}
\usepackage{lmodern}
\usepackage{times,latexsym}
\usepackage{url}
\usepackage[T1]{fontenc}

\usepackage[acceptedWithA]{tacl2018v2}

\usepackage{xspace,mfirstuc,tabulary}

\newif\iftaclinstructions
\taclinstructionsfalse 
\iftaclinstructions
\renewcommand{\confidential}{}
\renewcommand{\anonsubtext}{(No author info supplied here, for consistency with
TACL-submission anonymization requirements)}
\newcommand{\instr}
\fi

\iftaclpubformat 

\else

\fi

\usepackage[utf8x]{inputenc}
\usepackage{multirow}
\usepackage{wrapfig}
\usepackage{comment}
\usepackage{float}
\usepackage{bbold}
\usepackage{booktabs}
\usepackage{amsmath}
\usepackage{amssymb}
\usepackage{graphics}
\usepackage{graphicx}
\usepackage{longtable}

\usepackage{booktabs,makecell,tabularx}
\usepackage{array,booktabs}
\usepackage{siunitx}
\usepackage[obeyFinal,textsize=footnotesize]{todonotes}

\newcommand{\R}{\mathbb{R}}

\makeatletter
\newcommand{\@emptybiblabel}[1]{}
\makeatother

\title{Morphological analysis using a sequence decoder}

\author{Ekin Akyürek\Thanks{Equal contribution.} \\
\And
  Erenay Dayan{\i}k\footnotemark[1] \\
  Ko\c{c} University Artificial Intelligence Laboratory, 
  {\.I}stanbul, Turkey \\
{\tt eakyurek13,edayanik16,dyuret@ku.edu.tr}
 \\\And
  Deniz Yuret\Thanks{Corresponding author.} \\
  }

\date{}
\begin{document}
\maketitle
\begin{abstract}
We introduce Morse, 
a recurrent encoder-decoder model that produces morphological analyses of each word in a sentence.  The encoder turns the relevant information about the word and its context into a fixed size vector representation and the decoder generates the sequence of characters for the lemma followed by a sequence of individual morphological features. We show that generating morphological features individually rather than as a combined tag allows the model to handle rare or unseen tags and outperform whole-tag models. In addition, generating morphological features as a sequence rather than e.g.\ an unordered set allows our model to produce an arbitrary number of features that represent multiple inflectional groups in morphologically complex languages. We obtain state-of-the art results in nine languages of different morphological complexity under low-resource, high-resource and transfer learning settings. We also introduce TrMor2018, a new high accuracy Turkish morphology dataset. Our Morse implementation and the TrMor2018 dataset are available online to support future research\footnote{See \url{https://github.com/ai-ku/Morse.jl} for a Morse implementation in Julia/Knet \cite{knet2016mlsys} and \url{https://github.com/ai-ku/TrMor2018} for the new Turkish dataset.}.
\end{abstract}

\section{Introduction}

\begin{table}[ht]
\resizebox{0.5\textwidth}{!}{%
\begin{tabular}{|l|}
\hline
\textbf{Context \& analysis of ``masal{\i}''} \\ \hline
\hline
\textbf{masalı} yaz. (write \textbf{the tale}.) \\
masal+Noun+A3sg+Pnon+Acc\\  \hline
babamın \textbf{masalı} (my father's \textbf{tale}) \\
masal+Noun+A3sg+P3sg+Nom\\  \hline
mavi \textbf{masalı} oda (room \textbf{with a} blue \textbf{table}) \\
masa+Noun+A3sg+Pnon+Nom\textasciicircum{}DB+Adj+With\\ \hline
\end{tabular}%
}
\caption{Morphological analyses for Turkish word \textit{masal{\i}}. An example context and its translation is given before each analysis.}
\label{masali-table}
\end{table}

Morse is a recurrent encoder-decoder model that takes sentences in plain text as input and produces both lemmas and morphological features of each word as output. Table \ref{masali-table} presents an example: the ambiguous Turkish word ``\textit{masal{\i}}'' has three possible morphological analyses: the accusative and possessive forms of the stem ``\textit{masal}'' (tale) and the {\tt +With} form of the stem ``\textit{masa}'' (table), all expressed with the same surface form \cite{Oflazer1994a}. Morse attempts to output the correct analysis of each word based on its context in a sentence.

Accurate morphological analysis and disambiguation are important prerequisites for further syntactic and semantic processing, especially in morphologically complex languages. Many languages mark case, number, person etc.\ using morphology, which helps discover the correct syntactic dependencies. In agglutinative languages, syntactic dependencies can even be between subword units. For example \citet{Oflazer1999} observes that words in Turkish can have dependencies to any one of the inflectional groups of a derived word: in \textit{``mavi masalı oda''} (room with a blue table) the adjective \textit{``mavi''} (blue) modifies the noun root \textit{``masa''} (table) even though the final part of speech of ``masal\i'' is an adjective. This dependency would be difficult to represent without a detailed analysis of morphology.

We combined the following ideas to attack morphological analysis in the Morse model:
\begin{itemize}
\itemsep0em
\item Morse does not require an external rule based analyzer or dictionary, avoiding the parallel maintenance of multiple systems.
\item Morse performs lemmatization and tagging jointly by default, we also report on separating the two tasks.
\item Morse outputs morphological tags one feature at a time, giving it the ability to learn unseen/rare tags.
\item Morse generates features as a variable size seqeunce rather than a fixed set, allowing it to represent derivational morphology.
\end{itemize}

We evaluated our model on several Turkish datasets \cite{yuret2006learning,Yildiz:2016:MNM:3016100.3016302} and eight languages from the Universal Dependencies dataset \cite{Nivre2016UniversalDV} in low-resource, high-resource and transfer learning settings for comparison with existing work. We realized that existing Turkish datasets either had low inter-annotator agreement or small test sets, which made model comparison difficult due to noise and statistical significance problems. To address these issues we also created a new Turkish dataset, TrMor2018, which contains 460K tagged tokens and has been verified to be 96\% accurate by trained annotators. We report our results on this new dataset as well as previously available datasets.

The main contributions of this work are:
\begin{itemize} 
\itemsep0em 
\item A new encoder-decoder model that performs joint lemmatization and morphological tagging which can handle unknown words, unseen tag sequences, and multiple inflectional groups.
\item State-of-the-art results on nine languages of varying morphological complexity in low-resource, high-resource and transfer learning settings.
\item Release of a new morphology dataset for Turkish.
\end{itemize}

In the rest of the paper, we discuss related work in Section \ref{related}, detail our model’s input output representation and individual components in Section \ref{modelsec}, describe our datasets and introduce our new Turkish dataset in Section \ref{datasets}, present our experiments and results in Section \ref{exp setup}, and conclude in Section~\ref{conclusion}.
 
\section{Related Work}\label{related}

Morphological word analysis has been typically performed by solving multiple subproblems. In one common approach the subproblems of {\em lemmatization} (e.g.\  finding the stem ``masal'' for the first two examples in Table~\ref{masali-table} and ``masa'' for the third) and {\em morphological tagging} (e.g.\  producing {\tt +Noun+A3sg+Pnon+Acc} in the first example) are attacked separately. In another common approach a language-dependent rule based {\em morphological analyzer} outputs all possible lemma+tag analyses for a given word, and a statistical {\em disambiguator} picks the correct one in a given context. Even though Morse attacks these problems jointly, the prior work is best presented within these traditional divisions, contrasting various approaches with Morse where appropriate.

\subsection{Lemmatization and Tagging}

Early work in this area typically performed lemmatization and tagging separately. For example, the Shortest Edit Script (SES) approach to lemmatization classifies lemmas based on the mimimum sequence of operations that converts a wordform into a lemma \cite{Chrupala2006SimpleDC}. MarMoT \cite{mueller-schmid-schutze:2013:EMNLP} predicts the sequence of morphological tags in a sentence using a pruned higher order CRF.

SES was later extended to do joint lemmatization and morphological tagging in Morfette \cite{chrupala2008learning}, where two separate maximum entropy models are trained for predicting the lemma and the morphological tag and a third model returns a probability distribution over lemma-tag pairs. MarMoT was extended to Lemming \cite{muller-EtAl:2015:EMNLP}, which  used a joint log-linear model of lemmatization and tagging and provided empirical evidence that jointly modeling morphological tags and lemmata is mutually beneficial.

We chose to perform lemmatization and tagging jointly in Morse partly for linguistic reasons: as Table~\ref{masali-table} shows, a tag like {\tt +Noun+A3sg+Pnon+Acc} can be correct with respect to one lemma ({\tt masal}) and not another ({\tt masa}). For comparison with some of the earlier work, we did train Morse to only generate the morphological tag and observed some improvement in low-resource and transfer-learning settings but no significant improvement in high resource experiments.

More recent work started experimenting with deep learning models. \citet{heigold2017extensive} outperformed MarMoT in morphological tagging using a character based RNN encoder similar to Morse, combined with a whole-tag classifier. To address the data sparseness problem this work was extended in \citet{cotterell-heigold-2017-cross} with transfer learning, improving performance on low resource languages by up to 30\% using a related high resource language. 

\begin{table}[t]\centering
\begin{tabular}{|lrr|}
\hline
\textbf{Language} & \textbf{100sent} & \textbf{1000sent} \\ \hline \hline
Swedish & 9.19 & 1.02 \\
Bulgarian & 14.38 & 2.68 \\
Hungarian & 15.78 & 3.93 \\
Portuguese & 6.04 & 0.82 \\
\hline
\end{tabular}%
\caption{Percentage of tags in the test data that have been observed fewer than 5 times in the training data for four languages and two training sizes (100 and 1000 sentences).}
\label{unktable}
\end{table}

Morse uses a character based encoder that turns the relevant features of the word and its context into fixed size vector representations similar to \citet{heigold2017extensive}. Our main contribution is the {\em sequence decoder} that generates the characters of the lemma and/or morphological features sequentially one at a time. This is similar to the way rule based systems such as finite state transducers output morphological analyses. One advantage of generating features one at a time (e.g.\  {\tt +Acc}) rather than as a combined tag (e.g.\  {\tt +Noun+A3sg+Pnon+Acc}) is sample efficiency. Table~\ref{unktable} shows the percentage of tags in the test data that have been observed rarely in the training data for several languages. In low resource  experiments, we show that our sequence decoder significantly outperforms a variant that is trained to output full tags similar to \citet{heigold2017extensive} especially with unseen or rare tags.

\begin{figure}[t] 
    \centering
    \includegraphics[width=0.5\textwidth]{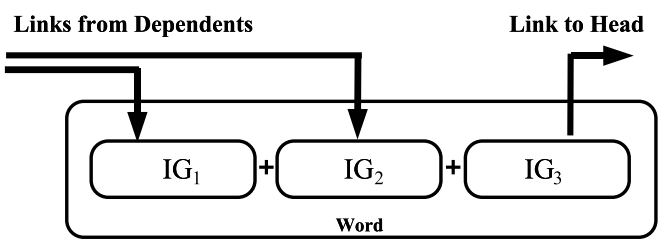}
\caption{Multiple inflectional groups in a word may have independent syntactic relationships. Figure from \cite{Eryigit06}.}
    \label{multiple-igs}
\end{figure}

\citet{malaviya2018neural} also avoids the data sparsity problem associated with whole-tags using a neural factor graph model to predict a set of features, improving the transfer learning performance. 
In contrast with \citet{malaviya2018neural}, Morse generates a variable number of features as a sequence rather than a fixed set. This allows it to adequately represent  derivations in morphologically complex words. For example, in the last analysis in Table~\ref{masali-table}, morphological features of the word ``\textit{masal\i}'' consist of two inflectional groups (IGs), a noun group and an adjective group, separated by a derivational boundary denoted by ``{\tt\textasciicircum{}DB}''. In ``\textit{mavi masalı oda}'' (room with a blue table) the adjective ``\textit{mavi}'' (blue) modifies the noun root ``\textit{masa}'' (table) even though the final part of speech of ``\textit{masal\i}'' is an adjective. In general, each IG in a morphologically complex word may have independent syntactic dependencies as shown in Figure~\ref{multiple-igs}. Thus for languages like Turkish, it is linguistically essential to be able to represent multiple IGs with a variable number of features \cite{J08-3003}. The sequence decoder approach of Morse outperforms the neural factor graph model of \citet{malaviya2018neural} in both low resource and transfer learning settings.

\subsection{Analysis and Disambiguation}
Morphological analysis is the task of producing all possible morphological parses for a given word. For morphologically simple languages like English, a dictionary is typically sufficient for this task \cite{baayen1995celex}. For morphologically complex languages like Turkish, the analysis can be performed by language dependent rule based systems such as finite-state transducers that encode morphophonemics and morphotactics \cite{koskenniemi1981application,koskenniemi1983two,karttunen1983two}. The first rule based analyzer for Turkish was developed in \citet{Oflazer1994a}, we used an updated version of this analyzer \cite{oflazer2018turkish2} when creating our new Turkish dataset. 

Morphological disambiguation systems take the possible parses for a given word from an analyzer and predict the correct one in a given context using 
rule based \cite{Karlsson:1995:CGL:546590,oflazer1994tagging,oflazer-tur-1996-combining,Daybelge07arule-based,daoud2009synchronized}, 
statistical \cite{hakkani2002statistical,yuret2006learning,spoustova-etal-2007-best}
or neural network based \cite{Yildiz:2016:MNM:3016100.3016302,DBLP:conf/coling/ShenCTLD16,toleu2017character}
techniques.
\citet{oflazer2018turkish3} provides a comprehensive summary for Turkish disambiguators.

Morse performs morphological analysis and disambiguation with a joint model partly to avoid using a separate morphological analyzer or dictionary. Having a single system combining morphological analysis and disambiguation is easier to use and maintain. The additional constraints brought by an external morphological analyzer or dictionary are certainly beneficial, but the benefit appears to be limited with sufficient data: In our experiments, (1) we outperform earlier systems that use separate morphological analysis and disambiguation components, and (2) when we use Morse only to disambiguate among the analyses generated by a rule-based analyzer, the accuracy gain is less than 1\% compared to generating analyses from scratch.

\section{Model} \label{modelsec}

\begin{figure*}[t] 
    \centering
    \includegraphics[width=\textwidth]{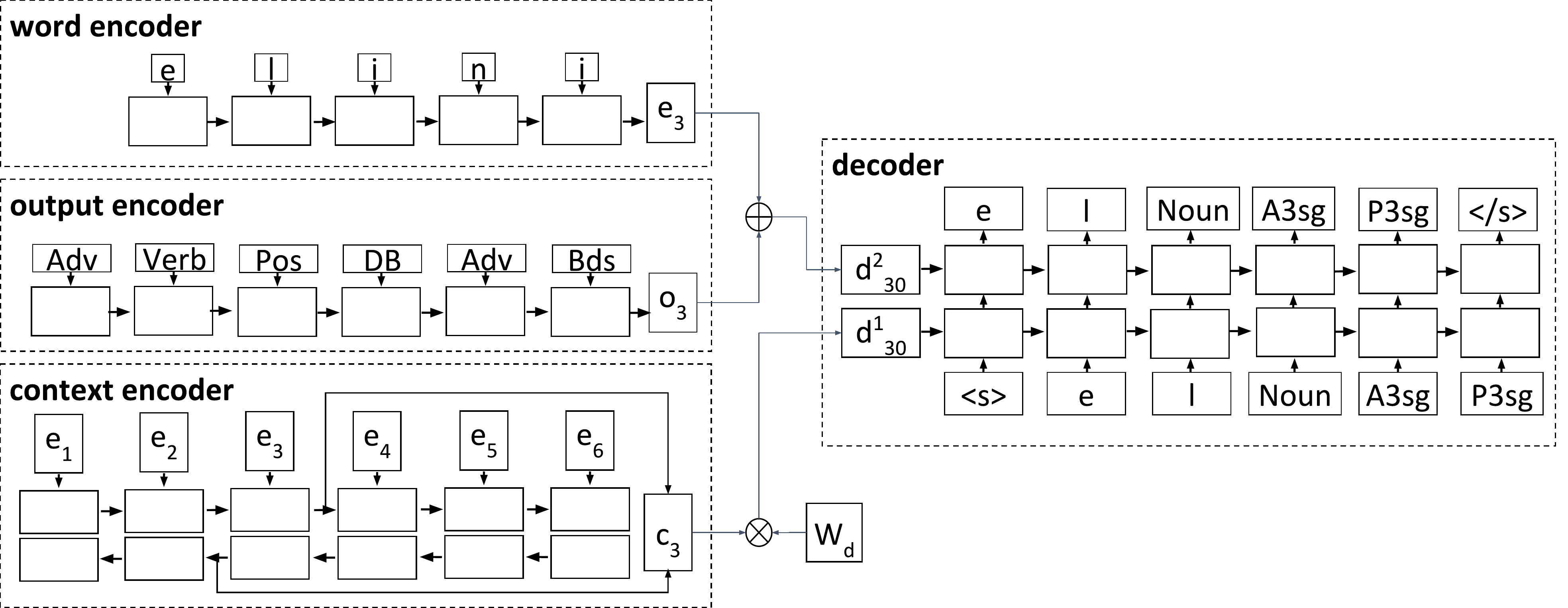}
\caption{Model illustration for the sentence \textit{"Sonra gülerek elini kardeşinin omzuna koydu"} (Then he laughed and put his hand on his brother's shoulder) and target word \textit{"elini"} (his hand). We use the morphological features of the words preceding the target as input to the output encoder: ``Sonra+Adv g\"ul+Verb+Pos\^{}DB+Adverb+ByDoingSo''.
}
    \label{fig:model}
\end{figure*}

Morse produces the morphological analysis (lemma plus morphological features) for each word in a given  sentence. It is loosely based on the sequence-to-sequence encoder-decoder network approach proposed by \citet{Sutskever:2014:SSL:2969033.2969173} for machine translation. However, we use three distinct encoders to create embeddings of various input features. First, a word encoder creates an embedding for each word based on its characters. Second, a context encoder creates an embedding for the context of each word based on the word embeddings of all words to the left and to the right.  Third, an output encoder creates an output embedding using the morphological features of the last two words.  These embeddings are fed to the decoder which produces the lemma and the morphological features of a target word one character/feature at a time. In the following subsections, we explain each component in detail.

\subsection{Input Output}
The input to the model consists of an $N$ word sentence $S=[w_1,\ldots,w_N]$, where $w_i$ is the i'th word in the sentence. Each word is input as a sequence of characters $w_i = [w_{i1},\ldots,w_{iL_i}], w_{ij}\in\mathcal{A}$ where $\mathcal{A}$ is the set of alphanumeric characters and $L_i$ is the number of characters in word $w_i$.

The output for each word consists of a lemma, a part-of-speech tag and a set of morphological features, e.g.\  
[m,~a,~s,~a,~l, Noun, A3sg, P3sg, Nom]
for ``masal\i''. The lemma is produced one character at a time, and the morphological information is produced one feature at a time. A sample output for a word looks like $[s_{i1},\ldots,s_{iR_i},f_{i1},\ldots,f_{iM_i}]$ where $s_{ij}\in\mathcal{A}$ is an alphanumeric character in the lemma, $R_i$ is the length of the lemma, $M_i$ is the number of features, $f_{ij}\in\mathcal{T}$ is a morphological feature from a feature set such as $\mathcal{T}=\{\mbox{Noun,Adj,Nom,A3sg,}\ldots\}$. 

We have experimented with other input-output formats as described in the experiments section: We found that jointly producing the lemma and the morphological features is more difficult than producing only morphological features in low-resource settings but gives similar performance in high-resource settings. We also found that generating the morphological tag one feature at a time rather than as a complete tag is advantageous, more so in morphologically complex languages and in low-resource settings.

\subsection{Word Encoder}
We map each character $w_{ij}$ to an $A$ dimensional character embedding vector $a_{ij}\in \R^A$.  The word encoder takes each word and processes the character embeddings from left to right producing hidden states $[h_{i1},\ldots,h_{iL_i}]$ where $h_{ij}\in\R^H$. The final hidden state $e_{i} = h_{iL_i}$ is used as the word embedding for word $w_i$. The top left box in Figure~\ref{fig:model} depicts the word encoder. We also experimented with external word embeddings but did not observe any significant improvement. 
\begin{eqnarray}\label{softmax}
  h_{ij} &=& \mbox{LSTM}(a_{ij},h_{ij-1}) \\
  h_{i0} &=& 0 \\
  e_i &=& h_{iL_i}
\end{eqnarray}

\subsection{Context Encoder} \label{context-encoder}
We use a bidirectional LSTM for the context encoder. The inputs are the word embeddings $e_{1},\cdots,e_{N}$ produced by the word encoder. The context encoder processes them in both directions and constructs a unique context embedding for each target word in the sentence. For a word $w_{i}$ we define its corresponding context embedding $c_{i}\in\R^{2H}$ as the concatenation of the forward $\overrightarrow{c}_i\in\R^H$ and the backward $\overleftarrow{c}_i\in\R^H$ hidden states that are produced after the forward and backward LSTMs process the word embedding $e_{i}$. The bottom left box in Figure~\ref{fig:model} depicts creation of the context vector for the target word ``elini''.
\begin{eqnarray}
		\overrightarrow{c}_{i} &=& \mbox{LSTM}_{f}(e_i,\overrightarrow{c}_{i-1}) \\
		\overleftarrow{c}_{i} &=&  \mbox{LSTM}_{b}(e_i,\overleftarrow{c}_{i+1}) \\
        \overrightarrow{c}_{0} &=& \overleftarrow{c}_{N+1} = 0 \\
        c_{i} &=& \left[ \overrightarrow{c}_i ; \overleftarrow{c}_{i}\right]
\end{eqnarray}
\subsection{Output Encoder} \label{output-encoder}
The output encoder captures information about the morphological features of words processed prior to each target word. For example, in order to assign the correct possessive marker to the word ``\textit{masal{\i}}'' (tale) in ``\textit{babam{\i}n masal{\i}}'' (my father's tale), it would be useful to know that the previous word ``\textit{babam{\i}n}'' (my father's) has a genitive marker.
During training we use the gold morphological features, during testing we use the output of the model. 


The output encoder only uses the morphological features, not the lemma characters, of the previous words as input: $[f_{11},\ldots,f_{1M_1},f_{21},\ldots,f_{i-1,M_{i-1}}]$. We map each morphological feature $f_{ij}$ to a $B$ dimensional feature embedding vector $b_{ij}\in\R^B$.  A unidirectional LSTM is run over the morphological features of the last two words to produce hidden states $[t_{11},\ldots,t_{i-1,M_{i-1}}]$ where $t_{ij}\in\R^H$. The final hidden state preceding the target word $o_i=t_{i-1,M_{i-1}}$ is used as the output embedding for word $w_i$. The middle left box in Figure~\ref{fig:model} depicts the output encoder.
\begin{eqnarray}\label{output-encoder-eq}
  t_{ij} &=& \mbox{LSTM}(b_{ij},t_{ij-1}) \\
  t_{i0} &=& t_{i-1,M_{i-1}} \\
  o_i &=& t_{i-1,M_{i-1}}
\end{eqnarray}

\begin{table*}[t]
\resizebox{\textwidth}{!}{%
\begin{tabular}{|crrrrrr|rrrrrrr|}
\hline
\textbf{lang} & \multicolumn{1}{c}{\textbf{train}} & \multicolumn{1}{c}{\textbf{dev}} & \multicolumn{1}{c}{\textbf{test}} & \multicolumn{1}{c}{\textbf{|T|}} & \multicolumn{1}{c}{\textbf{|F|}} & \multicolumn{1}{c|}{\textbf{|R|}} & \multicolumn{1}{c}{\textbf{lang}} & \multicolumn{1}{c}{\textbf{train}} & \multicolumn{1}{c}{\textbf{dev}} & \multicolumn{1}{c}{\textbf{test}} & \multicolumn{1}{c}{\textbf{|T|}} & \multicolumn{1}{c}{\textbf{|F|}} & \multicolumn{1}{c|}{\textbf{|R|}}\\ \hline \hline
DA & 80378 & 10332 & 10023 & 159 & 44 & 0.03\% & SV & 66645 & 9797 & 20377 & 211 & 40 & 0.06\% \\
RU & 75964 & 11877 & 11548 & 734 & 39 & 0.27\% & BG & 124336 & 16089 & 15724 & 439 & 45 & 0.03\% \\
FI & 162621 & 18290 & 21041 & 2243 & 93 & 0.68\% & HU & 20166 & 11418 & 10448 & 716 & 73 & 1.03\% \\
ES & 384554 & 37349 & 12069 & 404 & 46 & 0.03\% & PT & 211820 & 11158 & 10468 & 380 & 47 & 0.03\% \\ \hline
\end{tabular}
}
\caption{Data statistics of UD Version 2.1 Treebanks. The values in the \{train, dev, test\} columns are the number of tokens in the splits. $|T|$ gives the number of distinct tags (pos~+ morphological features), 
$|F|$ the number of distinct feature values. |R| gives the unseen tag percentage in the test set.}
\label{data-UD}
\end{table*}

\subsection{Decoder}
The decoder is implemented as a 2-Layer LSTM network that outputs the correct lemma+tag for a single target word\footnote{We also experimented with two variants of our model: MorseTag only outputs morphological features, and MorseDisamb uses the decoder to rank probabilities of a set of analyses provided by a rule-based system.}. By conditioning on the three encoder embeddings and its own hidden state, the decoder learns to generate $y_{i} = [y_{i1},\ldots,y_{iK_i}]$  where $y_{i}$ is the correct sequence for the target word $w_{i}$ in sentence $S$, $y_{ij} \in \mathcal{A} \cup \mathcal{T}$ represents both lemma characters and morphological feature tokens, and $K_i$ is the total number of output tokens (lemma + features) for word $w_i$. The first layer of the decoder is initialized with the context embedding $c_{i}$.
\begin{eqnarray}\label{context-eq}
		d_{i0}^1 &=& \mbox{relu}(W_d \times c_{i} \oplus W_{db}) \\
        d_{ij}^1 &=& \mbox{LSTM}(y_{ij-1},d_{ij-1}^1)
\end{eqnarray}
\noindent where, $W_d \in \mathbb{R}^{H\times 2H}$, $W_{db} \in \mathbb{R}^{H}$ and $\oplus$ is element-wise summation. We initialize the second layer with the word and output embeddings after combining them by element-wise summation. 
\begin{eqnarray}\label{hs1-eq}
	d_{i0}^2 &=& e_{i} + o_{i} \\
    d_{ij}^2 &=& \mbox{LSTM}(d_{ij}^1,d_{ij-1}^2)
\end{eqnarray}

We parameterize the distribution over possible morphological features and characters at each time step as 
\begin{equation}\label{softmax-eq}
	\small p(y_{ij}|d_{ij}^2) = \mbox{softmax}(W_s \times d_{ij}^2 \oplus W_{sb})
\end{equation}
where  $W_s \in \mathbb{R}^{|\mathcal{Y}|\times H }$  and $W_{sb} \in \mathbb{R}^{|\mathcal{Y}|}$ where $\mathcal{Y}=\mathcal{A}\cup\mathcal{T}$ is the set of characters and morphological features in output vocabulary. The right side of Figure~\ref{fig:model} depicts the decoder.

\section{Datasets}\label{datasets}
We evaluate Morse on several different languages and datasets. First we describe the multilingual datasets we used from the Universal Dependency Datasets \cite{Nivre2016UniversalDV}.  We then describe two existing datasets for Turkish and introduce our new dataset TrMor2018.

\subsection{Universal Dependency Datasets}
We tested Morse on eight languages selected from the Universal Dependency Datasets Version 2.1 \cite{Nivre2016UniversalDV}. In Table~\ref{data-UD}, we summarize the corpus statistics. Specifically, we use the CoNLL-U format\footnote{\url{http://universaldependencies.org/format.html}} for the input files, take column 2 (FORM) as input and predict columns 3 (LEMMA), 4 (UPOSTAG) and 6 (FEATS). We show the number of distinct features with $|F|$, the number of distinct composite tags with $|T|$, and the unseen composite tag percentage with $|R|$ to indicate the morphological complexity of a language.

\subsection{Turkish Datasets}

\begin{table}[t]
\centering
\resizebox{0.5\textwidth}{!}{%
\begin{tabular}{@{}|llll|@{}}
\hline
\textbf{Dataset} & \textbf{Ambig} & \textbf{Unamb} & \textbf{Total} \\ \hline 
\hline
TrMor2006Train & 398290   & 439234  & 837524 \\
TrMor2006Test  & 379      & 483     & 862    \\
TrMor2016Test  & 9460     & 9802    & 19262  \\
TrMor2018 & 216803 & 243866 & 460669 \\
\hline
\end{tabular}
}
\caption{Number of ambiguous, unambiguous and all tokens for datasets TrMor2006 \cite{yuret2006learning}, TrMor2016 \cite{Yildiz:2016:MNM:3016100.3016302} (which shares the same training set), and TrMor2018 (introduced in this paper).}
\label{table-stat-tr}
\end{table}

For Turkish we evaluate our model on three datasets described in Table~\ref{table-stat-tr}. These datasets contain derivational as well as inflectional morphology represented by multiple inflectional groups as described in the Introduction. In contrast, the UD datasets only preserve  information in the last inflectional group.

The first dataset, TrMor2006, was provided by Kemal Oflazer and published in \citet{yuret2006learning} based on a Turkish newspaper dataset. The training set was disambiguated semi-automatically and has limited accuracy. The test set was hand-tagged but is very small (862 tokens) to reliably distinguish between models with similar accuracy. We randomly extracted 100 sentences from the training set and used them as the development set while training our model.

\begin{table*}[t]
\centering
\resizebox{\textwidth}{!}{%
\begin{tabular}{|cl|lllll|}
\hline
\textbf{HR/LR} & 
\textbf{Model} &
\textbf{LR100} & 
\textbf{XFER100} & 
\textbf{LR1000} &
\textbf{XFER1000} & 
\textbf{HR} 
\\ \hline \hline
\multirow{4}{*}{DA/SV}
	& Cotterell	& 15.11	& 66.06	& 68.64	& 82.26	& 91.79 \\ 
	& Malaviya	& 29.47	& 63.22	& 71.32	& 77.43	& 	\\ 
	& Morse	& 62.45(0.69)	& 72.70(0.59)	& 86.44(0.17)	& 87.55(0.22)	& 92.68(0.19)	\\ 
	& MorseTag	& \textbf{66.19(1.23)}	& \textbf{76.70(0.72)}	& \textbf{88.31(0.17)}	& \textbf{88.97(0.54)}	& \textbf{93.35(0.23)}	\\ 
	\hline
\multirow{4}{*}{RU/BG}
	& Cotterell	& 29.05	& 52.76	& 59.20	& 71.90	& 82.02	\\ 
	& Malaviya	& 27.81	& 46.89	& 39.25	& 67.56	& 	\\ 
	& Morse	& 59.82(1.65)	& 69.27(0.54)	& 87.71(0.26)	& 88.70(0.16)	& 85.43(0.12)	\\ 
	& MorseTag	& \textbf{66.97(1.34)}	& \textbf{75.78(0.26)}	& \textbf{88.96(0.41)}	& \textbf{90.52(0.21)}	& \textbf{86.51(0.36)}	\\ 
	\hline
\multirow{4}{*}{FI/HU}
	& Cotterell	& 21.97	& 51.74	& 50.75	& 61.80	& 85.25	\\ 
	& Malaviya	& 33.32	& 45.41	& 45.90	& 63.93	& 	\\ 
	& Morse	& 49.58(1.27)	& 54.84(0.71)	& \textbf{72.28(0.74)}	& 71.33(1.83)	& \textbf{91.24(0.28)}	\\ 
	& MorseTag	& \textbf{54.87(0.72)}	& \textbf{57.12(0.36)}	& \textbf{73.55(0.72)}	& \textbf{73.86(1.28)}	& \textbf{91.42(0.84)}	\\ 
	\hline
\multirow{4}{*}{ES/PT}
	& Cotterell	& 18.91	& 79.40	& 74.22	& 85.85	& \textbf{93.09}	\\ 
	& Malaviya	& 58.82	& 77.75	& 76.26	& 85.02	& 	\\ 
	& Morse	& \textbf{70.57(0.54)}	& 80.01(0.38)	& \textbf{86.29(0.31)}	& 87.51(0.27)	& \textbf{92.95(0.21)}	\\ 
	& MorseTag	& \textbf{70.80(1.14)}	& \textbf{81.60(0.16)}	& \textbf{86.24(0.28)}	& \textbf{88.01(0.13)}	& \textbf{92.89(0.18)}	\\ 
 \hline
 \end{tabular}
 }
 \caption{Accuracy comparisons for UDv2.1 Datasets. Table~\ref{UDResultsF1} gives F1 comparisons which are similar. LR is the low resource language, HR is the high resource language, XFER represents HR to LR transfer learning. 100/1000 indicate the number of sentences in the training set for low resource experiments. Morse and MorseTag rows give the average of 5 experiments with standard deviation in parentheses. Statistically significant leaders ($p<0.05$) are marked in bold. Some experiments have multiple leaders marked when their difference is not statistically significant.}
\label{UDResultsAcc}
 \end{table*}
 
\begin{table*}[t]
\centering
\resizebox{\textwidth}{!}{%
\begin{tabular}{|cl|lllll|}

 \hline
\textbf{HR/LR} & 
\textbf{Model} &
\textbf{LR100} & 
\textbf{XFER100} & 
\textbf{LR1000} &
\textbf{XFER1000} & 
\textbf{HR} 
\\ \hline \hline
\multirow{4}{*}{DA/SV}
	& Cotterell	& 08.36	& 73.95	& 76.36	& 87.88	& 94.18 \\
	& Malaviya	& 54.09	& 78.75	& 84.42	& 87.56	&  \\
	& Morse	& 72.77(0.74)	& 81.39(0.27)	& 91.52(0.07)	& 92.42(0.15)	& 95.18(0.11) \\
	& MorseTag	& \textbf{74.91(1.26)}	& \textbf{84.27(0.48)}	& \textbf{92.39(0.26)}	& \textbf{93.04(0.35)}	& \textbf{95.50(0.21)} \\
	\hline
\multirow{4}{*}{RU/BG}
	& Cotterell	& 14.32	& 58.41	& 67.22	& 77.89	& 90.63 \\
	& Malaviya	& 40.97	& 64.46	& 60.23	& 82.06	&  \\
	& Morse	& 68.90(1.36)	& 76.86(0.41)	& 92.38(0.13)	& 93.12(0.21)	& 93.08(0.03) \\
	& MorseTag	& \textbf{75.52(1.16)}	& \textbf{83.60(0.06)}	& \textbf{93.08(0.37)}	& \textbf{94.24(0.11)}	& \textbf{93.55(0.13)} \\
	\hline
\multirow{4}{*}{FI/HU}
	& Cotterell	& 13.30	& 68.15	& 58.68	& 75.96	& 90.54 \\
	& Malaviya	& 54.88	& 68.63	& 74.05	& 85.06	&  \\
	& Morse	& 65.17(1.17)	& 71.77(0.42)	& 85.96(0.42)	& 85.91(0.86)	& \textbf{95.34(0.20)} \\
	& MorseTag	& \textbf{72.21(0.67)}	& \textbf{74.17(0.14)}	& \textbf{87.17(0.38)}	& \textbf{87.39(0.53)}	& \textbf{95.37(0.52)} \\
	\hline
\multirow{4}{*}{ES/PT}
	& Cotterell	& 07.10	& 86.03	& 81.62	& 91.91	& \textbf{96.57} \\
	& Malaviya	& 73.67	& 88.42	& 87.13	& 92.35	&  \\
	& Morse	& \textbf{80.06(0.73)}	& 88.11(0.25)	& \textbf{92.43(0.28)}	& 93.31(0.20)	& \textbf{96.52(0.10)} \\
	& MorseTag	& \textbf{80.07(0.92)}	& \textbf{88.99(0.42)}	& \textbf{92.29(0.28)}	& \textbf{93.56(0.14)}	& \textbf{96.44(0.13)} \\
 \hline
\end{tabular}
}
\caption{F1 comparisons for UDv2.1 Datasets. See Table~\ref{UDResultsAcc} for column descriptions.}
\label{UDResultsF1}
\end{table*}

The second dataset, TrMor2016, was prepared by \citet{Yildiz:2016:MNM:3016100.3016302}. The training set is the same with TrMor2006 but they manually retagged a subset of the training set containing roughly 20000 tokens to be used as a larger test set. Unfortunately they did not exclude the sentences in the test set from the training set in their experiments.  Futhermore, they do not provide any inter-annotator agreement results on the new test set.

Given the problems associated with these datasets, we decided to prepare a new dataset, TrMor2018, that we release with this paper. Our goal is to provide a dataset with high inter-annotator agreement that is large enough to allow dev/test sets of sufficient size to distinguish model performances in a statistically significant manner.  The new dataset consists of 34673 sentences and 460669 tokens in total from different genres of Turkish text.

TrMor2018 was annotated semi-automatically in multiple passes. The initial pass was performed automatically by a previous state-of-the-art model \cite{yuret2006learning}. The resulting data was spot checked in multiple passes for mistakes and inconsistencies by annotators, prioritizing ambiguous high frequency words. Any systematic errors discovered were corrected by hand written scripts.

In order to monitor our progress, we randomly selected a subset and disambiguated all of it manually. This subset contains 2090 sentences and 26819 words. Two annotators annotated each word independently and we assigned the final morphological tag of each word based on the adjudication by a third. Taking this hand-tagged subset as the gold standard, we measure the noise level in the corresponding semi-automatic results after every pass. Importantly, the hand-tagged subset is only used for evaluating the noise level of the main dataset, i.e. we do not use it for training or testing, and we do not use the identity of the mistakes to inform our passes. Our current release of TrMor2018 has a disagreement level of 4.4\% with the hand-tagged subset, which is the current state-of-the-art for Turkish morphological datasets.

\section{Experiments and Results} \label{exp setup}

In this section we describe our training procedure, give experimental results, compare with related models, and provide an ablation analysis. The results demonstrate that Morse, generating analyses with its sequence decoder, significantly outperforms the state of the art in low resource, high resource and transfer learning experiments. We also experimented with two variants of our model for more direct comparisons: MorseTag which only predicts tags without lemmas, and MorseDisamb which chooses among the analyses generated by a rule-based morphological analyzer.

\subsection{Training}
All LSTM units have $H=512$ hidden units in our experiments. The size of the character embedding vectors are $A=64$ in the word encoder. In the decoder part, the size of the output embedding vectors are $B=256$. We initialized model parameters with Xavier initialization \cite{glorot2010understanding}. 

Our networks are trained using back-propagation through time with stochastic gradient descent. The learning rate is set to $lr{=}1.6$ and is decayed based on the development accuracy. We apply learning rate decay by a factor of 0.8 if the development set accuracy is not improved after 5 consecutive epochs. Likewise, early-stopping is forced if the development set accuracy is not improved after 10 consecutive epochs, returning the model with the best dev accuracy. To reduce overfitting, dropout is applied with the rates of 0.5 for low-resource and 0.3 in high-resource settings for each of the LSTM units as well as embedding layers.

\begin{table*}[t]
\centering
\begin{tabular}{|lccc|}
\hline
\textbf{Method} & \textbf{TrMor2006} & \textbf{TrMor2016} & \textbf{TrMor2018} \\
\hline
\cite{yuret2006learning}  & 95.82  & - & - \\
\cite{Sak2007} & 96.28   & -  & - \\
\cite{Yildiz:2016:MNM:3016100.3016302}  & - & 92.20 & - \\
\cite{DBLP:conf/coling/ShenCTLD16} & 96.41 & - & - \\
Morse  & 95.94   & 92.63 & 97.67 \\ 
MorseDisamb & 96.52 & \textbf{92.82} & \textbf{98.59} \\
\hline
\end{tabular}
\caption{Test set lambda+tag accuracy of several models on Turkish Datasets: TrMor2006 \cite{yuret2006learning}, TrMor2016 \cite{Yildiz:2016:MNM:3016100.3016302}, TrMor2018 (published with this paper).
}
\label{tr-results}
\end{table*}

\subsection{Multilingual Results}
\label{UDResults}

For comparison with existing work, we evaluated our model on four pairs of high/low resource language pairs: Danish/Swedish (DA/SV), Russian/Bulgarian (RU/BG), Finnish/Hungarian (FI/HU), Spanish/Portuguese (ES/PT). Table~\ref{UDResultsAcc} compares the accuracy and Table~\ref{UDResultsF1} compares the F1 scores of four related models\footnote{Accuracy is for the whole-tag ignoring the lemma. The F1 score is based on the precision and recall of each morphological feature ignoring the lemma, similar to \citet{malaviya2018neural}.}: (1) Cotterell: a classification based model with a similar encoder that predicts whole tags rather than individual features \cite{cotterell-heigold-2017-cross}, (2) Malaviya: a neural factor graph model that predicts a fixed number of morphological features rather than variable length feature sequences \cite{malaviya2018neural}, (3) Morse: our model with joint prediction of the lemma and the tag (the lemma is ignored in scoring), and (4) MorseTag: a version of our model that predicts only the morphological tag without the lemma (Cotterell and Malaviya only predict tags). We compare results in three different settings: (1) LR100 and LR1000 columns show the low resource setting where we experiment with 100 and 1000 sentences of training data in Swedish, Bulgarian, Hungarian and Portuguese, (2) XFER100 and XFER1000 columns show the transfer learning setting where the related high resource language is used to help improve the results of the low resource language which has only 100/1000 sentences, and (3) HR column gives the high resource setting where we use the full training data with the high resource languages Danish, Russian, Finnish and Spanish\footnote{Malaviya is missing from the HR column because we could not train it with large datasets in a reasonable amount of time. For Cotterell we used the SPECIFIC model given in \citet{malaviya2018neural} in all experiments.}.

For transfer experiments we use a simple transfer scheme: training with the high-resource language for 10 epochs and using the resulting model to initialize the compatible weights of the model for the low-resource language. All LSTM weights and embeddings for identical tokens are transferred exactly, new token embeddings are initialized randomly.

In all low resource, transfer learning and high resource experiments, Morse and MorseTag perform significantly better than the two related models (with the single exception of the high resource experiment on Spanish, a morphologically simple language, where the difference with Cotterell is not statistically significant). This supports the hypothesis that the sequence decoder of Morse is more sample efficient compared to a whole-tag model or a neural factor graph model.

Tag-only prediction in MorseTag generally outperforms joint lemma-tag prediction in Morse but the difference decreases or disappears with more training data and in simpler languages. In half of the high resource experiments, their difference is not statistically significant. The difference is also insignificant in most of the experiments with the morphologically simplest language pair Spanish/Portuguese.

\subsection{Turkish Results}
\label{TRResults}

Table-\ref{tr-results} shows the lemma+tag test accuracy of several systems for different Turkish datasets. We masked digits and \texttt{Prop} (proper noun) tags in our evaluations. The older models use a hand-built morphological analyzer \cite{Oflazer1994a} that gives a list of possible lemma+tag analyses and train a disambiguator to pick the correct one in the given context. Standard Morse works without a list of analyses, the decoder can generate the lemma+tag from scratch. Older disambiguators always get 100\% accuracy on unambiguous tokens with a single analysis, whereas Morse may fail to generate the correct lemma+tag pair. In order to make a fair comparison we also tested a version of Morse that disambiguates among a given set of analyses by comparing the probability assigned to them by the decoder (MorseDisamb).

MorseDisamb gives the best results across all three datasets. The best scores are printed in bold where the difference is statistically significant. None of the differences in TrMor2006 are statistically significant because of the small size of the test set. In TrMor2016 both Morse and MorseDisamb give state of the art results. The TrMor2018 results were obtained using an average of 5 random splits into 80\%, 10\%, 10\% for training, validation and test sets. 

Note that the numbers for the three datasets are significantly different. Each result naturally reflects the remaining errors and biases in the corresponding dataset, which might result
in the true accuracy figure being higher or lower. In spite of these imperfections, we believe the new TrMor2018 dataset will allow for better comparison of different models in terms of learning efficiency thanks to its larger size and lower noise level.


\begin{table}[t]
\centering
\resizebox{0.5\textwidth}{!}{%
\begin{tabular}{|lccc|}
\hline
\textbf{Method}  & \textbf{A} & \textbf{U} & \textbf{T}  \\ \hline
\hline
word                    & 94.38          & 98.70          & 96.72         \\
word+context            & 96.21          & 98.52          & 97.69         \\
word+context+output     & 96.43          & 98.80          & 97.79        \\ \hline
\end{tabular}}
\caption{Ablation analysis test set performances on the TrMor2018 dataset. A: \textbf{A}mbiguous Accuracy,  U: \textbf{U}nambiguous accuracy T: \textbf{T}otal accuracy}
\label{results-ablation}
\end{table}

\subsection{Ablation Analysis}
In this section, the contributions of the individual components of the full model are analyzed. In the following three ablation studies, we disassemble or change individual modules to investigate the change in the performance of the model. We use the TrMor2018 dataset in the first two experiments and UD Datasets in the last experiment. Table~\ref{results-ablation} presents the results.


\begin{table*}[t]
\centering
\begin{tabular}{|crccrccrcc|}
\hline
 & 
\multicolumn{3}{c}{\textbf{count=0}} & \multicolumn{3}{c}{\textbf{count\textless{}100}} & \multicolumn{3}{c|}{\textbf{count$\geq$100}} \\ 
\textbf{Lang} & \textbf{Tok} & \textbf{Tag} & \textbf{Seq} & \textbf{Tok} & \textbf{Tag} & \textbf{Seq} & \textbf{Tok} & \textbf{Tag} & \textbf{Seq}  \\ \hline \hline
SV & 12 & 0.0 & 8.33 &   844 & 81.28 & 82.82 &         19521 & 94.49 & 94.65 \\
BG & 4 & 0.0 & 0.0 &     910 & 81.32 & 83.41 &         14810 & 96.62 & 97.37 \\
HU & 108 & 0.0 & 20.37 &   2333 & 53.54 & 59.24 &        8007 & 78.24 & 80.67 \\
PT & 3 & 0.0 & 0.0 &     207 & 63.29 & 67.63 &         9991 & 93.04 & 92.25 \\ \bottomrule
\end{tabular}
\caption{Test accuracy for tags that were observed 0, $< 100$ and $\geq 100$ times in the 1000 sentence training sets. \textbf{Tok} is the number of tokens with the specified count, \textbf{Tag} is the accuracy using a whole-tag classifier, \textbf{Seq} is the accuracy using a sequence decoder.}
\label{DecoderAblation}
\end{table*}

\begin{table*}[t]
\centering
\begin{tabular}{|lrcrcrc|}
\hline
& \multicolumn{2}{c}{\textbf{count=0}} & \multicolumn{2}{c}{\textbf{count\textless{}5}} & \multicolumn{2}{c|}{\textbf{count$\geq$5}} \\ 
\textbf{Dataset} & \textbf{Tok} & \textbf{Acc} & \textbf{Tok} & \textbf{Acc} & \textbf{Tok} & \textbf{Acc}  \\ \hline \hline

TRMor2006 & 30 & 86.67 &   16 & 100.0 & 816 & 98.9 \\
TRMor2016 & 79 & 2.53 &   579 & 93.78 &   18570 & 98.48\\
TRMor2018 &  0 & - &     1702 & 82.78 &  45119 & 99.48\\
UD-DA & 1019 & 71.84 &	 1023 & 94.72 &	 7981 & 98.93\\
UD-ES & 593 & 79.26 &     627 & 95.37 &	 10780 & 99.36\\
UD-FI & 2279 & 61.34 &	 1802 & 88.85 &	 16989 & 98.21\\
UD-RU & 1656 & 77.48 &	 1587 & 94.39 &	 8305 & 99.22\\ \bottomrule


\end{tabular}
\caption{Test accuracy for lemmas that were observed 0, $< 5$ and $\geq 5$ times in the TRMor and UD datasets. \textbf{Tok} is the number of tokens with the specified count, \textbf{Acc} is the accuracy using Morse.}
\label{LemmaGeneration}
\end{table*}

We start our ablation studies by  removing both the context encoder and the output encoder leaving only the word encoder. The resulting model (word) is a standard sequence-to-sequence model which only uses the characters in the target word without any context information. This gives us a baseline and shows that more than 95\% of the wordforms can be correctly tagged ignoring the context.

We then improve the model by adding the context encoder (word+context). We observe a 1.83\% increase in ambiguous word accuracy and 0.97\% in overall accuracy. This version is capable of learning more than only a single morphologic analysis of each wordform. As an example, the  lemma ``\textit{röportaj}'' (interview) has 5 distinct wordforms observed in the training set. We tested both models on the never before seen wordform  ``röportajı'' in ``\textit{Benden bu röportajı yalanlamamı rica etti.}'' (I was asked to deny the interview). While (word) failed by selecting the most frequently occurring tag of ``röportaj'' in the training set ({\tt Noun+A3sg+Pnon+Nom}), word+context disambiguated the target wordform successfully ({\tt +Noun+A3sg+Pnon+Acc}), demonstrating the ability to generalize to unseen wordforms.

Finally, we add the output encoder to reconstruct the full Morse model (word+context+output). We observe a further 0.22\% increase in ambiguous word accuracy and 0.10\% increase in overall accuracy.  These experiments show that each of the model components have a positive contribution to the overall performance.

We believe our ablation models have several advantages over a standard sequence-to-sequence model: Both the input and the output of the system needs to be partly character based to analyze morphology and to output lemmas. This leads to long input and output sequences. By running the decoder separately for each word, we avoid the necessity to squeeze the information in the whole input sequence into a single vector. A standard sequence-to-sequence model would also be more difficult to evaluate as it may produce zero or multiple outputs for a single input token or produce outputs that are out of order. A per-word decoder avoids these alignment problems as well.

To compare our approach to whole-tag classifiers like \citet{heigold2017extensive}, we created two versions of the (word+context) model, one with a sequence decoder and one with a whole-tag classifier. We trained these models on Turkish and UD datasets to test unseen/rare tag and lemma generation. Table~\ref{DecoderAblation} shows the accuracy of each model on three sets of tags: unseen tags, tags that were seen less than 100 times and tags that were seen at least 100 times in the training set. The sequence decoder generally performs better across different frequency ranges. In particular, results confirm that the sequence decoder can generate some unseen tags correctly while the whole-tag classifier in principle cannot. We observe that the advantage is smaller for more frequent tags, in fact the whole-tag classifier performs better with the most frequent tags in Portuguese, a morphologically simple language. A similar trend is observed in Table~\ref{LemmaGeneration} for lemma generation: Morse is able to generate a significant percent of the unseen/rare lemmas correctly.

\section{Conclusion}
\label{conclusion}
In this paper, we presented Morse, a language independent character based encoder-decoder architecture for morphological analysis and TrMor2018, a new Turkish morphology dataset manually confirmed to have 96\% inter-annotator agreement. The Morse encoder employs two different unidirectional LSTMs to obtain word and output embeddings and a bidirectional LSTM to obtain the context embedding of a target word. The Morse decoder outputs the lemma of the word one character at a time followed by the morphological tag, one feature at a time. We evaluated Morse on nine different languages, and obtained state-of-the art results on all of them. We provided empirical evidence that producing morphological features as a sequence outperforms methods that produce whole tags or feature sets, and the advantage is more significant in low resource settings. 

To our knowledge, Morse is the first deep learning model that performs joint lemmatization and tagging, that performs well with unknown and rare wordforms and tags, and that can produce a variable number of features in multiple inflectional groups to represent derivations in morphologically complex languages.

\section*{Acknowledgments}
We would like to thank Kemal Oflazer and all student annotators for their help in creating the TrMor2018 dataset, the editors and anonymous reviewers for their many helpful comments. This work was supported by the Scientific and Technological Research Council of Turkey (T\"{U}B\.{I}TAK) grants 114E628 and 215E201.

\bibliography{tacl}

\begin{thebibliography}{33}
\expandafter\ifx\csname natexlab\endcsname\relax\def\natexlab#1{#1}\fi

\bibitem[{Baayen et~al.(1995)Baayen, Piepenbrock, and
  Gulikers}]{baayen1995celex}
R~Harald Baayen, Richard Piepenbrock, and Leon Gulikers. 1995.
\newblock The {CELEX} lexical database (release 2).
\newblock \emph{Distributed by the Linguistic Data Consortium, University of
  Pennsylvania}.

\bibitem[{Chrupala(2006)}]{Chrupala2006SimpleDC}
Grzegorz Chrupala. 2006.
\newblock Simple data-driven context-sensitive lemmatization.
\newblock \emph{Procesamiento del Lenguaje Natural}, 37.

\bibitem[{Chrupala et~al.(2008)Chrupala, Dinu, and van
  Genabith}]{chrupala2008learning}
Grzegorz Chrupala, Georgiana Dinu, and Josef van Genabith. 2008.
\newblock Learning morphology with {Morfette}.
\newblock \emph{LREC 2008}.

\bibitem[{Cotterell and Heigold(2017)}]{cotterell-heigold-2017-cross}
Ryan Cotterell and Georg Heigold. 2017.
\newblock \href {https://doi.org/10.18653/v1/D17-1078} {Cross-lingual
  character-level neural morphological tagging}.
\newblock In \emph{Proceedings of the 2017 Conference on Empirical Methods in
  Natural Language Processing}, pages 748--759, Copenhagen, Denmark.
  Association for Computational Linguistics.

\bibitem[{Daoud(2009)}]{daoud2009synchronized}
Daoud Daoud. 2009.
\newblock Synchronized morphological and syntactic disambiguation for {Arabic}.
\newblock \emph{Advances in Computational Linguistics}, 41:73--86.

\bibitem[{Daybelge and \c{C}i\c{c}ekli(2007)}]{Daybelge07arule-based}
Turhan Daybelge and \.{I}lyas \c{C}i\c{c}ekli. 2007.
\newblock A rule-based morphological disambiguator for {Turkish}.
\newblock In \emph{Proceedings of Recent Advances in Natural Language
  Processing (RANLP 2007), Borovets}, pages 145--149.

\bibitem[{Eryi{\u{g}}it et~al.(2008)Eryi{\u{g}}it, Nivre, and
  Oflazer}]{J08-3003}
G{\"u}l{\c{s}}en Eryi{\u{g}}it, Joakim Nivre, and Kemal Oflazer. 2008.
\newblock \href {https://doi.org/10.1162/coli.2008.07-017-R1-06-83} {Dependency
  parsing of {Turkish}}.
\newblock \emph{Computational Linguistics}, 34(3).

\bibitem[{Eryi{\u{g}}it and Oflazer(2006)}]{Eryigit06}
G{\"u}l{\c{s}}en Eryi{\u{g}}it and Kemal Oflazer. 2006.
\newblock \href {http://aclweb.org/anthology/E/E06/E06-1012.pdf} {Statistical
  dependency parsing of {Turkish}}.
\newblock In \emph{Proceedings of the 11th EACL}, pages 89--96, Trento.

\bibitem[{Glorot and Bengio(2010)}]{glorot2010understanding}
Xavier Glorot and Yoshua Bengio. 2010.
\newblock Understanding the difficulty of training deep feedforward neural
  networks.
\newblock In \emph{Proceedings of the Thirteenth International Conference on
  Artificial Intelligence and Statistics}, pages 249--256.

\bibitem[{Haji{\v{c}} et~al.(2007)Haji{\v{c}}, Votrubec, Krbec,
  Kv{\v{e}}to{\v{n}} et~al.}]{hajivc2007best}
Jan Haji{\v{c}}, Jan Votrubec, Pavel Krbec, Pavel Kv{\v{e}}to{\v{n}}, et~al.
  2007.
\newblock The best of two worlds: Cooperation of statistical and rule-based
  taggers for {Czech}.
\newblock In \emph{Proceedings of the Workshop on Balto-Slavonic Natural
  Language Processing: Information Extraction and Enabling Technologies}, pages
  67--74. Association for Computational Linguistics.

\bibitem[{Hakkani-T{\"u}r et~al.(2002)Hakkani-T{\"u}r, Oflazer, and
  T{\"u}r}]{hakkani2002statistical}
Dilek~Zeynep Hakkani-T{\"u}r, Kemal Oflazer, and G{\"o}khan T{\"u}r. 2002.
\newblock Statistical morphological disambiguation for agglutinative languages.
\newblock \emph{Computers and the Humanities}, 36(4):381--410.

\bibitem[{Hakkani-Tür et~al.(2018)Hakkani-Tür, Sara\c{c}lar, Tür, Oflazer,
  and Yuret}]{oflazer2018turkish3}
Dilek~Zeynep Hakkani-Tür, Murat Sara\c{c}lar, Gökhan Tür, Kemal Oflazer, and
  Deniz Yuret. 2018.
\newblock \href {https://books.google.com/books?id=D-5lDwAAQBAJ} {Morphological
  disambiguation for {T}urkish}.
\newblock In K.~Oflazer and M.~Sara{\c{c}}lar, editors, \emph{Turkish Natural
  Language Processing}, Theory and Applications of Natural Language Processing,
  chapter~3. Springer International Publishing.

\bibitem[{Heigold et~al.(2017)Heigold, Neumann, and van
  Genabith}]{heigold2017extensive}
Georg Heigold, G{\"u}nter Neumann, and Josef van Genabith. 2017.
\newblock An extensive empirical evaluation of character-based morphological
  tagging for 14 languages.
\newblock In \emph{Proceedings of the 15th Conference of the European Chapter
  of the Association for Computational Linguistics}, volume~1, pages 505--513.

\bibitem[{Karlsson et~al.(1995)Karlsson, Voutilainen, Heikkila, and
  Anttila}]{Karlsson:1995:CGL:546590}
Fred Karlsson, Atro Voutilainen, Juha Heikkila, and Arto Anttila, editors.
  1995.
\newblock \emph{Constraint Grammar: A Language-Independent System for Parsing
  Unrestricted Text}.
\newblock Walter de Gruyter \& Co., Hawthorne, NJ, USA.

\bibitem[{Karttunen and Wittenburg(1983)}]{karttunen1983two}
Lauri Karttunen and Kent Wittenburg. 1983.
\newblock A two-level morphological analysis of {English}.
\newblock In \emph{Texas Linguistic Forum Austin, Tex.}, 22, pages 217--228.

\bibitem[{Koskenniemi(1981)}]{koskenniemi1981application}
Kimmo Koskenniemi. 1981.
\newblock An application of the two-level model to {Finnish}.
\newblock \emph{Computational morphosyntax: Report on research}, 1984:19--41.

\bibitem[{Koskenniemi(1983)}]{koskenniemi1983two}
Kimmo Koskenniemi. 1983.
\newblock Two-level model for morphological analysis.
\newblock In \emph{IJCAI}, volume~83, pages 683--685.

\bibitem[{Malaviya et~al.(2018)Malaviya, Gormley, and
  Neubig}]{malaviya2018neural}
Chaitanya Malaviya, Matthew~R. Gormley, and Graham Neubig. 2018.
\newblock \href {http://aclweb.org/anthology/P18-1247} {Neural factor graph
  models for cross-lingual morphological tagging}.
\newblock In \emph{Proceedings of the 56th Annual Meeting of the Association
  for Computational Linguistics (Volume 1: Long Papers)}, pages 2653--2663.
  Association for Computational Linguistics.

\bibitem[{Mueller et~al.(2013)Mueller, Schmid, and
  Sch\"{u}tze}]{mueller-schmid-schutze:2013:EMNLP}
Thomas Mueller, Helmut Schmid, and Hinrich Sch\"{u}tze. 2013.
\newblock \href {http://www.aclweb.org/anthology/D13-1032} {Efficient
  higher-order {CRFs} for morphological tagging}.
\newblock In \emph{Proceedings of the 2013 Conference on Empirical Methods in
  Natural Language Processing}, pages 322--332, Seattle, Washington, USA.
  Association for Computational Linguistics.

\bibitem[{M\"{u}ller et~al.(2015)M\"{u}ller, Cotterell, Fraser, and
  Sch\"{u}tze}]{muller-EtAl:2015:EMNLP}
Thomas M\"{u}ller, Ryan Cotterell, Alexander Fraser, and Hinrich Sch\"{u}tze.
  2015.
\newblock \href {http://aclweb.org/anthology/D15-1272} {Joint lemmatization and
  morphological tagging with lemming}.
\newblock In \emph{Proceedings of the 2015 Conference on Empirical Methods in
  Natural Language Processing}, pages 2268--2274, Lisbon, Portugal. Association
  for Computational Linguistics.

\bibitem[{Nivre et~al.(2016)Nivre, de~Marneffe, Ginter, Goldberg, Hajic,
  Manning, McDonald, Petrov, Pyysalo, Silveira, Tsarfaty, and
  Zeman}]{Nivre2016UniversalDV}
Joakim Nivre, Marie-Catherine de~Marneffe, Filip Ginter, Yoav Goldberg, Jan
  Hajic, Christopher~D. Manning, Ryan~T. McDonald, Slav Petrov, Sampo Pyysalo,
  Natalia Silveira, Reut Tsarfaty, and Daniel Zeman. 2016.
\newblock Universal dependencies v1: A multilingual treebank collection.
\newblock In \emph{LREC}.

\bibitem[{Oflazer(1994)}]{Oflazer1994a}
Kemal Oflazer. 1994.
\newblock \href {/ref/oflazer/BU-CEIS-9304.pdf} {Two-level description of
  {Turkish} morphology}.
\newblock \emph{Literary and Linguistic Computing}, 9(2):137--148.

\bibitem[{Oflazer(2018)}]{oflazer2018turkish2}
Kemal Oflazer. 2018.
\newblock \href {https://books.google.com/books?id=D-5lDwAAQBAJ} {Morphological
  processing for {T}urkish}.
\newblock In K.~Oflazer and M.~Sara{\c{c}}lar, editors, \emph{Turkish Natural
  Language Processing}, Theory and Applications of Natural Language Processing,
  chapter~2. Springer International Publishing.

\bibitem[{Oflazer et~al.(1999)Oflazer, Hakkani-T{\"u}r, and
  T{\"u}r}]{Oflazer1999}
Kemal Oflazer, Dilek~Zeynep Hakkani-T{\"u}r, and G{\"o}khan T{\"u}r. 1999.
\newblock \href {/ref/oflazer/ttbank.pdf} {Design for a {Turkish} treebank}.
\newblock In \emph{Proceedings of the Workshop on Linguistically Interpreted
  Corpora, EACL 99}, Bergen, Norway.

\bibitem[{Oflazer and Kuru{\"o}z(1994)}]{oflazer1994tagging}
Kemal Oflazer and {\.I}lker Kuru{\"o}z. 1994.
\newblock Tagging and morphological disambiguation of {Turkish} text.
\newblock In \emph{Proceedings of the Fourth Conference on Applied Natural
  Language Processing}, pages 144--149. Association for Computational
  Linguistics.

\bibitem[{Oflazer and T{\"u}r(1996)}]{oflazer-tur-1996-combining}
Kemal Oflazer and Gokhan T{\"u}r. 1996.
\newblock \href {https://www.aclweb.org/anthology/W96-0207} {Combining
  hand-crafted rules and unsupervised learning in constraint-based
  morphological disambiguation}.
\newblock In \emph{Conference on Empirical Methods in Natural Language
  Processing}.

\bibitem[{Sak et~al.(2007)Sak, G{\"u}ng{\"o}r, and Sara{\c{c}}lar}]{Sak2007}
Ha{\c{s}}im Sak, Tunga G{\"u}ng{\"o}r, and Murat Sara{\c{c}}lar. 2007.
\newblock \href {https://doi.org/10.1007/978-3-540-70939-8_10} {Morphological
  disambiguation of {Turkish} text with perceptron algorithm}.
\newblock In Alexander Gelbukh, editor, \emph{Computational Linguistics and
  Intelligent Text Processing: 8th International Conference, CICLing 2007},
  pages 107--118. Springer Berlin Heidelberg, Berlin, Heidelberg.

\bibitem[{Shen et~al.(2016)Shen, Clothiaux, Tagtow, Littell, and
  Dyer}]{DBLP:conf/coling/ShenCTLD16}
Qinlan Shen, Daniel Clothiaux, Emily Tagtow, Patrick Littell, and Chris Dyer.
  2016.
\newblock The role of context in neural morphological disambiguation.
\newblock In \emph{{COLING}}, pages 181--191. {ACL}.

\bibitem[{Sutskever et~al.(2014)Sutskever, Vinyals, and
  Le}]{Sutskever:2014:SSL:2969033.2969173}
Ilya Sutskever, Oriol Vinyals, and Quoc~V. Le. 2014.
\newblock \href {http://dl.acm.org/citation.cfm?id=2969033.2969173} {Sequence
  to sequence learning with neural networks}.
\newblock In \emph{Proceedings of the 27th International Conference on Neural
  Information Processing Systems}, NIPS'14, pages 3104--3112, Cambridge, MA,
  USA. MIT Press.

\bibitem[{Toleu et~al.(2017)Toleu, Tolegen, and
  Makazhanov}]{toleu2017character}
Alymzhan Toleu, Gulmira Tolegen, and Aibek Makazhanov. 2017.
\newblock Character-aware neural morphological disambiguation.
\newblock In \emph{Proceedings of the 55th Annual Meeting of the Association
  for Computational Linguistics (Volume 2: Short Papers)}, volume~2, pages
  666--671.

\bibitem[{Y{\i}ld{\i}z et~al.(2016)Y{\i}ld{\i}z, Tirkaz, \c{S}ahin, Eren, and
  S\"onmez}]{Yildiz:2016:MNM:3016100.3016302}
Eray Y{\i}ld{\i}z, \c{C}a\u{g}lar Tirkaz, H.~Bahad{\i}r \c{S}ahin,
  Mustafa~Tolga Eren, and Ozan S\"onmez. 2016.
\newblock \href {http://dl.acm.org/citation.cfm?id=3016100.3016302} {A
  morphology-aware network for morphological disambiguation}.
\newblock In \emph{Proceedings of the Thirtieth AAAI Conference on Artificial
  Intelligence}, AAAI'16, pages 2863--2869. AAAI Press.

\bibitem[{Yuret(2016)}]{knet2016mlsys}
Deniz Yuret. 2016.
\newblock Knet: beginning deep learning with 100 lines of julia.
\newblock In \emph{Machine Learning Systems Workshop at NIPS 2016}.

\bibitem[{Yuret and T{\"u}re(2006)}]{yuret2006learning}
Deniz Yuret and Ferhan T{\"u}re. 2006.
\newblock Learning morphological disambiguation rules for {Turkish}.
\newblock In \emph{Proceedings of the main conference on Human Language
  Technology Conference of the North American Chapter of the Association of
  Computational Linguistics}, pages 328--334. Association for Computational
  Linguistics.

\end{thebibliography}
\bibliographystyle{acl_natbib}

\end{document}